\documentclass[10pt,twocolumn,letterpaper]{article}

\usepackage[pagenumbers]{cvpr} 

\usepackage{times}
\usepackage{epsfig}
\usepackage{graphicx}
\usepackage{amsmath}
\usepackage{amssymb}
\usepackage{subfiles}
\usepackage{subcaption}

\usepackage{comment}
\usepackage{boldline} 
\usepackage{dsfont}
\usepackage{soul}
\usepackage{multirow}
\usepackage{xspace}

\newcommand{\fsor}{FSOR\xspace}
\newcommand{\fsl}{FSL\xspace}
\newcommand{\gfsl}{GFSL\xspace}
\newcommand{\gfsor}{GFSOR\xspace}
\newcommand{\closet}{closed-set\xspace}

\newcommand{\spm}{\scriptstyle \pm}

\newcommand{\img}[1]{\mathcal{D}^{#1}}
\newcommand{\supp}[1]{\mathcal{S}_{#1}}
\newcommand{\query}[1]{\mathcal{Q}_{#1}}
\newcommand{\class}[1]{\mathcal{C}^{#1}}
\newcommand{\task}[1]{\mathcal{T}^{#1}}
\newcommand{\proto}[1]{\mathbf{p}_{#1}}
\newcommand{\protos}[1]{\mathbf{P}^{#1}}
\newcommand{\backbone}{{f}}
\newcommand{\gen}{{f_n}}

\newcommand{\att}[1]{\mathbf{A}_{(#1)}}
\newcommand{\attkernel}[1]{\mathbf{K}_{#1}}

\newcommand{\semanvec}[1]{\mathbf{e}_{#1}}
\newcommand{\semanvis}[1]{\mathbf{z}_{#1}}
\newcommand{\gating}{{f_g}}
\newcommand{\thre}[1]{\theta_{#1}}

\usepackage[pagebackref,breaklinks,colorlinks]{hyperref}

\usepackage[capitalize]{cleveref}
\crefname{section}{Sec.}{Secs.}
\Crefname{section}{Section}{Sections}
\Crefname{table}{Table}{Tables}
\crefname{table}{Tab.}{Tabs.}

\def\cvprPaperID{9328} 
\def\confName{CVPR}
\def\confYear{2022}

\begin{document}

\title{Task-Adaptive Negative Envision for Few-Shot Open-Set Recognition}

\author{Shiyuan Huang$^{*}$ \qquad \qquad Jiawei Ma$^{*}$  \qquad \qquad Guangxing Han  \qquad\qquad Shih-Fu Chang \\
		Columbia University \\ {\tt\small \{shiyuan.h, jiawei.m, gh2561, sc250\}@columbia.edu }}
\maketitle

\begin{abstract}
We study the problem of few-shot open-set recognition (FSOR), which learns a recognition system capable of both fast adaptation to new classes with limited labeled examples and rejection of unknown negative samples. Traditional large-scale open-set methods have been shown ineffective for FSOR problem due to data limitation. Current FSOR methods typically calibrate few-shot closed-set classifiers to be sensitive to negative samples so that they can be rejected via thresholding. However, threshold tuning is a challenging process as different FSOR tasks may require different rejection powers. In this paper, we instead propose task-adaptive negative class envision for FSOR to integrate threshold tuning into the learning process. Specifically, we augment the few-shot closed-set classifier with additional negative prototypes generated from few-shot examples. By incorporating few-shot class correlations in the negative generation process, we are able to learn dynamic rejection boundaries for FSOR tasks. Besides, we extend our method to generalized few-shot open-set recognition (GFSOR), which requires classification on both many-shot and few-shot classes as well as rejection of negative samples. Extensive experiments on public benchmarks validate our methods on both problems. \footnote{Code available at \url{https://github.com/shiyuanh/TANE}}
\end{abstract}


\section{Introduction}

With the emergence of large-scale image datasets \cite{deng2009imagenet, lin2014microsoft, everingham2010pascal}, deep learning has achieved great success in various vision tasks \cite{russakovsky2015imagenet, redmon2016you, cuturi2013sinkhorn, R_RPN, han2018semi, SSD_TDR}. Current recognition systems usually assume a predefined set of classes with sufficient number of labeled data. Each testing sample is supposed to belong to these predefined classes so that the systems only need to perform closed-set classification.

\begin{figure}[t]
\begin{center}
 \includegraphics[width=1.0\linewidth]{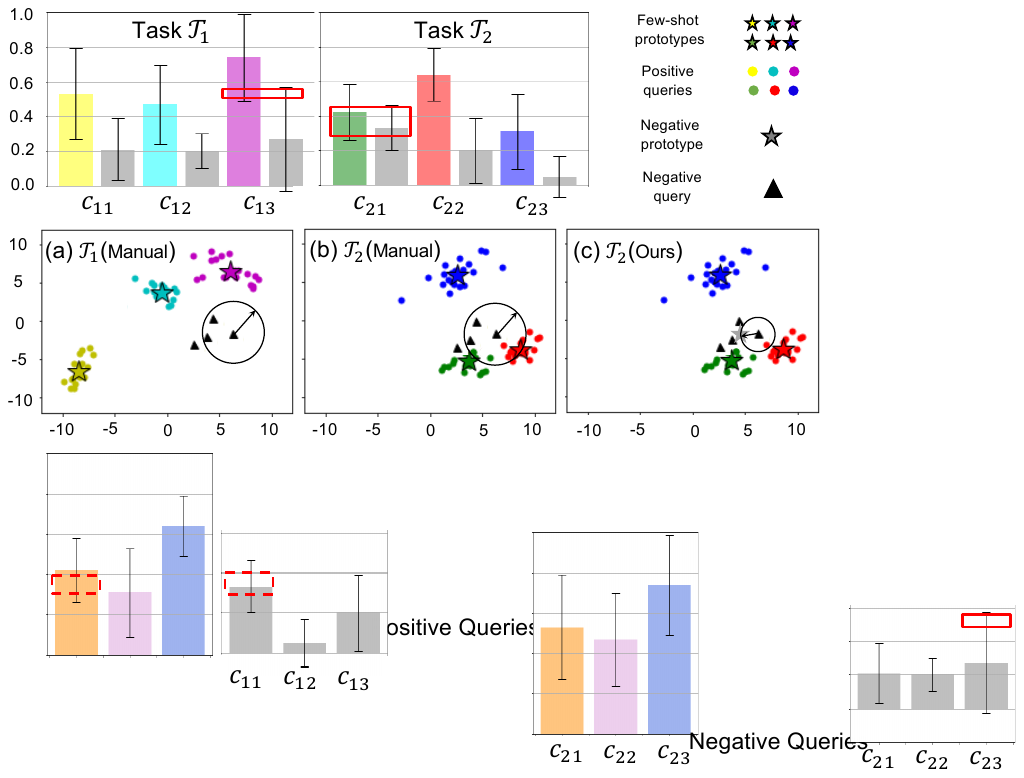}
\end{center}
 \caption{
 (Up): For each few-shot class in few-shot open-set recognition (\fsor) task $\task{}_1$ and $\task{}_2$, we calculate the detection scores (similarity) of both negative and positive queries and find their mean and standard deviation. (We use the same negative queries for both tasks.) However, a negative query may have similar detection score to the positive queries (highlighted by red box).
 (Down): In addition, to reject a negative  query, existing \fsor method relies on a manually selected threshold. However, a rejection threshold working properly for  $\task{}_1$ may fail in  $\task{}_2$. Instead, we propose to 
 learn a negative prototype that automatically estimates a task-adaptive threshold for negative detection.
 }
\label{fig:concept}
\end{figure}

In real-world applications, we face more challenging recognition scenarios. First, sufficient labeled training data are hardly guaranteed due to high cost of data collection and possibly limited access to sensitive or rare data. Few-shot learning~\cite{finn2017model, sung2018learning, vinyals2016matching, snell2017prototypical} (FSL) typically tackles data insufficiency scenario by  fast adaptation of recognition system to new classes with access to very few (e.g., only one) labeled instances. But FSL still holds a \closet assumption. 

On the other hand, there are efforts that endow a recognition system with the ability to handle out-of-distribution testing samples. 
Open-set recognition (OR)~\cite{scheirer2013open, openmax2016, schlachter2019open, goodfellow2014generative, mirza2014conditional} considers the case where testing samples could come from other unknown source under a large-scale training setting. Without loss of capability of classifying closed-set queries (i.e., positive queries), it also needs to detect queries from unknown classes (i.e., negative queries). Current OR methods typically learns an open-set classifier by either calibrating prediction scores or synthesizing negative queries. They rely on large amount of data to avoid overfitting and estimate distribution properly. But with only a few labeled instances, it becomes hard to do so. Hence direct application of OR methods under few-shot setting degrades the performance significantly\cite{Liu_2020_CVPR, jeong2021few}.

We aim to develop a model solving for both challenges, i.e., few-shot open-set recognition (FSOR). The goal of \fsor is to both 1) accept \& recognize \textit{positive queries} from few-shot classes with very few labeled samples and 2) detect \textit{negative queries} from undisclosed (\textit{negative}) classes. Previous FSOR methods ~\cite{Liu_2020_CVPR, jeong2021few} provide meta-learning-based solutions for learning threshold-based negative detector. 
They calibrate few-shot close-set classifier and output a rejection score for each testing sample. A sample is rejected if the rejection score is above a certain rejection threshold, which has to be manually defined. 
However, as shown in Fig.~\ref{fig:concept}, a good recognition performance relies heavily on a good choice of threshold: (a) a few-shot classifier may have similar detection score for a negative query and a positive query, where different thresholds need to be set separately; (b) to reject a negative query, a threshold works properly for one task may fail in other tasks. In summary, threshold tuning could be a challenging process as different FSOR tasks contain different few-shot classes that may need very different rejection powers to determine outliers.

In this paper, we instead propose to integrate threshold tuning into the learning process for \fsor. We extend the few-shot classifier with additional prototypes that represent the negative class. Specifically, a negative  generator is applied on few-shot class prototypes and learns negative prototypes across tasks via meta-learning, so that negative prototypes can serve as task-adaptive rejection boundaries for different \fsor tasks. A testing query is then rejected if the prediction scores on all few-shot classes are lower than that on the negative prototype. 
We study the design of negative generator and experimentally demonstrate an optimal solution that involves task-level information into the negative prototype envision.
We also introduce the concept of conjugate task of \fsor where two \fsor tasks are considered conjugate if the few-shot class in one task can be used to simulate unknown sources in the other.  To this end, we propose a \textit{conjugate training}  strategy to facilitate the learning process. Moreover, we consider a new but more challenging problem, generalized FSOR (\gfsor), where the recognition system needs to classify on both many-shot and few-shot classes as well as reject negative samples. In this case, negative prototypes are generated from both many-shot and few-shot classes. 
We name our method of learning negative prototypes as \textit{task-adaptive negative class envision}.

Our method is validated by extensive experiments on public benchmarks for both \fsor and \gfsor problems. In summary, our contributions are as follows: 
\begin{enumerate}
	\item We provide a threshold-free solution for few-shot open-set recognition (\fsor), where we extend classifier with negative prototypes that compute task-adaptive rejection boundaries. 
	
	\item We provide a study of the negative prototype generator design and experimentally demonstrate an optimal solution involving task-level knowledge for negative envision. In addition, we propose an efficient and novel training strategy, \textit{conjugate training}, to facilitate the learning process. 
	
	\item Extensively evaluated on public benchmarks, our approach is able to achieve SOTA performance. We further formulate the problem of generalized FSOR where our method is also shown to be effective. 
\end{enumerate}

In the following sections, we will discuss related literature in \fsl, OR and \fsor (Sec.~\ref{sec:related_works}); In Sec.~\ref{sec:formulation} we formally define \fsor and \gfsor tasks  and go over existing threshold-based meta-learning solutions. In Sec.~\ref{sec:approach} we present our approach of task-adaptive  negative envision. Finally in Sec.~\ref{sec:experiment} we demonstrate the experimental analysis and results of our approach.

\section{Related Works}\label{sec:related_works}
\noindent \textbf{Few-Shot Learning.} \fsl aims for fast adaptation to new recognition task with very few labeled examples. Meta-learning is widely used to learn transferable knowledge upon a set of tasks using episodic training. There are mainly two types of meta-learning approaches: 
1). Optimization-based method \cite{finn2017model, nichol2018first, flennerhag2019meta, sun2019meta} modifies the gradient back-propagation so that the parameter updates can be more sensitive to the few training examples; 2). Metric-based methods \cite{vinyals2016matching, snell2017prototypical, sung2018learning, NEURIPS2018_66808e32,Han_2022_AAAI,Han_2022_CVPR,Han_2021_ICCV, ypsilantis2021met} learns to obtain an optimal metric space so that a class with the highest similarity is assigned to the query. 
As an extension on FSL, 
generalized \fsl (\gfsl) \cite{gidaris2018dynamic} learns to expand many-shot classifier with novel classes using a few training data. 
Both (G)\fsl  hold a closed-set assumption where testing queries belong to novel classes (or many-shot classes in \gfsl). Our work instead extends (G)\fsl to open-set setting. 
\\

\noindent \textbf{Large-Scale Open-Set Recognition.} 
OR aims to learn a classifier sensitive to negative queries that come from unknown classes.
OR methods typically include class probability re-calibration~\cite{openmax2016, schlachter2019open, lee2017training} and negative sample synthesis with generative methods~\cite{ge2017generative,Neal_2018_ECCV}. Those methods typically assume large number of training data. A most relevant work to us \cite{zhou2021learning} also proposes to augment classifier to learn adaptive rejection thresholds. But it relies on large-scale data to train the augmented classifier from scratch, while ours generates negative prototypes based on few-shot classes. Direct application of OR methods to few-shot setting fails or degrades the performance \cite{Liu_2020_CVPR, jeong2021few} mainly due to over-fitting. Our work instead provides a few-shot-specific OR solution to deal with limited data. 
\\

\noindent \textbf{Few-Shot Open-Set Recognition.}  To bridge \fsl and OR, recently \cite{Liu_2020_CVPR} provides a meta-learning-based solution for \fsor that introduces an open-set loss in the meta-training process to calibrate few-shot prototype-based classifier. 
\cite{jeong2021few} improves the limitation of negative sampling in \cite{Liu_2020_CVPR} by imposing a transformation consistency regularization on few-shot samples. However, their methods are threshold-based, which require careful selection of thresholds to perform good recognition. Instead, we propose a threshold-free solution to overcome the challenge.  
\\

\begin{figure}[t]
\begin{center}
 \includegraphics[width=1.0\linewidth]{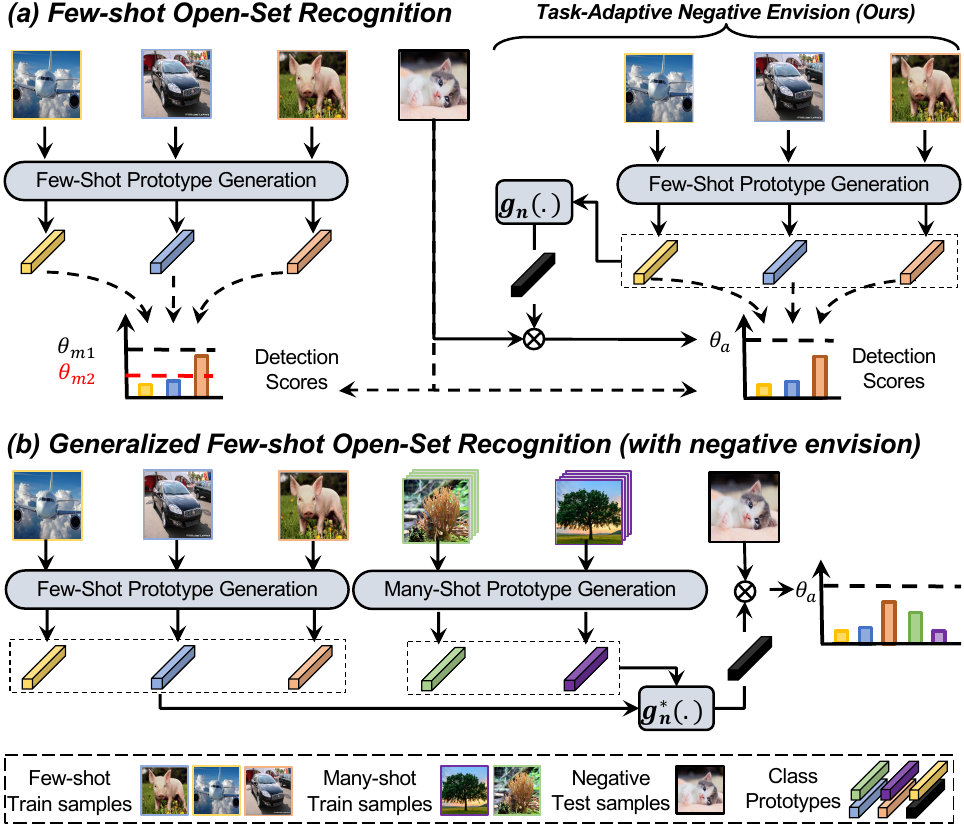}
\end{center}
 \caption{
 (a) Comparison between conventional threshold-based \fsor methods and ours with negative envision. Conventional methods reject a sample if the detection scores of all classes are below a  carefully selected threshold ($\theta_{m1}$). An improperly selected threshold ($\theta_{m2}$), on the other hand, would result in recognition failure. Instead, we propose to envision a negative prototype to \emph{learn to} estimate threshold ($\theta_{a}$) for each instance dynamically within the task. (b) For a GFSOR task, the negative prototype is generated from both many-shot and few-shot prototypes.
 }
\label{fig:comparison}
\end{figure}

\section{Problem Formulation}\label{sec:formulation}

With only a few labeled training samples, few-shot open-set recognition (FSOR) aims to 1) detect negative queries that come from \textit{unknown} sources and 2) correctly classify positive queries. 
Formally, a FSOR task can be denoted as $\task{} = (\supp{}, \query{}^f, \query{}^n | \class{f})$ where $\class{f}$ refer to the few-shot classes that have  few labeled training samples (also called supports): $\supp{}=\cup_{c\in\class{f}}\supp{c}$. The goal is to learn a recognition model with the supports so that during testing time, it can successfully classify positive queries $\query{}^f$ and detect  negative queries $\query{}^n$. We denote $\query{} = \query{}^f \cup \query{}^n$ as the entire query set.
We call a FSOR task \textit{$N$-way $K$-shot} if we have $|\class{f}|=N$ and $|\supp{c}|=K$ for all $c \in \class{f}$. Briefly speaking,  the only difference between \fsor and conventional \fsl tasks is that \fsor has additional negative queries that need to be rejected.

Existing \fsor approaches~\cite{Liu_2020_CVPR, jeong2021few} are built upon the popular metric-based \fsl method ProtoNet\cite{snell2017prototypical}, and our approach also follows the same fashion. Below we provide more context on ProtoNet. 

ProtoNet\cite{snell2017prototypical} learns a prototype-based few-shot classifier. In detail, each few-shot class $c \in \class{f}$ is represented by a prototype 
$\proto{c}$, computed by the average of $K$ support features: $\proto{c} = \frac 1 K \sum_{s \in \supp{c}} \backbone(s)$, where $\backbone$ is a feature extractor and $\backbone(s) \in \mathcal{R}^d$.
Then, all prototypes $\protos{f} = \{\proto{c}\}_{c\in\class{f}}$ build up a closed-set classifier where a positive query $q \in \query{}^f$ can be classified by nearest neighborhood search, \ie,
\begin{equation}
    \text{argmax}_{c}\big(\{f_s(f(q),\proto{c})\}_{c \in \class{f}} \big),
\end{equation}
where $f_s(\cdot,\cdot)$ is a function to measure the closeness between two inputs, \eg, cosine similarity. 

In order to learn an open-set classifier, existing \fsor approaches \cite{Liu_2020_CVPR, jeong2021few} calibrate the few-shot close-set classifier to get per-class detection scores and reject via thresholding. 
As illustrated in Fig.~\ref{fig:comparison}(a), for threshold-based \fsor methods, a threshold $\thre{m}$ needs to be manually set and a negative query $q \in \query{}^n$ will be rejected if all of the detection scores are below $\thre{m}$, \ie, $\text{max}\big(\{f_s(f(q),\proto{c})\}_{c \in \class{f}} \big) < \thre{m}$.

In addition to \fsor tasks, we further consider a more realistic situation where both few-shot classes $\class{f}$ and many-shot classes $\class{*}$ (i.e., containing large amount of labeled data) exist, resulting an imbalanced distribution. To this end, we  formulate the generalized few-shot open-set recognition (GFSOR) task $\task{*} = (\supp{}, \query{}^{*}, \query{}^f, \query{}^n | \class{*},\class{f})$ where $\query{}^{*}$ are queries from $\class{*}$. And the goal is to correctly classify both $\query{}^{*} \cup \query{}^f$ and reject negative query $\query{}^{n}$. Similarly, we call a \gfsor task \textit{$N$-way $K$-shot} if we have $|\class{f}|=N$ and $|\supp{c}|=K$ for all $c \in \class{f}$.

\section{Approach} \label{sec:approach}
Here we present our threshold-free approach towards (G)\fsor. We first provide an overview of how to use negative envision to estimate task-adaptive rejection boundaries; then we provide a list of negative generators used in practice; finally we introduce conjugate training which encourages the learning process from task mutual supervision.  

\subsection{Overview}
Fig.~\ref{fig:comparison} provides an overview of our \textit{Task-Adaptive Negative Envision} approach and how it compares to threshold-based methods. Threshold-based methods ~\cite{Liu_2020_CVPR, jeong2021few} calculate per-class detection scores and manually define a threshold for rejection. Without a carefully cherry-picked a threshold for each task, it's hard to successfully detect  $q \in \query{}^n$ across different tasks (Fig.~\ref{fig:comparison}(a)).
Instead, we expand classifier with negative prototype $\proto{}^{-}$ that are computed from few-shot class prototypes $\protos{f}$ via a negative generator $g_n(\cdot)$. When a query comes in, it's able to automatically calculate a task-specific threshold from the negative prototype:
\begin{equation}\label{eq:avg}
    \thre{a} = f_s(f(q),\proto{}^{-}).
\end{equation}
Then, a negative query $q\in \query{}^n$ will be rejected if $\text{max}\big(\{f_s(f(q),\proto{c})\}_{c \in \class{f}} \big) < \thre{a}$.
As such, rejection boundaries are dynamically estimated with respect to few-shot classes $\class{f}$ and support instances $\supp{}$. Our approach can be also applied to \gfsor tasks where the negative prototype are generated from both few-shot and many-shot class prototypes to get task-adaptive threshold with respect to both few-shot and many-shot classes.

\subsection{Negative Generator}
To find the best negative generator $g_n(\cdot)$, we explore different choices which we describe below in detail.
\subsubsection{MLP}
We start with a simple generator that consists of a single MLP layer applied on averaged class prototypes, i.e.,  
\begin{equation}
    \proto{}^{-} = \gen(\proto{avg}^{-}),~~\proto{avg}^{-} = \frac{1}{N}\sum\nolimits_{c\in\class{f}}\proto{c}
\end{equation}
where $\gen$ is a MLP that takes $\proto{avg}^{-}$, the mean of $\protos{f}$, as input so that $\proto{}^-$ is independent from the prototype order. 
Meanwhile, we set $\proto{avg}^{-}$ as a naive baseline (\textbf{AVG}) as $\proto{avg}^{-}$ is also order-independent.

\subsubsection{ATT}
Transformers\cite{vaswani2017attention} are proved effective in exploiting relations, which is also independent from the input order (without positional encoding). 
We apply a standard Transformer attention block over few-shot class prototypes to generate the negative prototype. 
Specifically, we calculate the self-attention weight between class prototypes, i.e., 
\begin{equation*}
\att{\protos{f},\protos{f}} = \frac 1 {\sqrt{d}} (\protos{f}\attkernel{q} (\protos{f}\attkernel{k})^T),
\end{equation*}
where $\att{\protos{f},\protos{f}} \in \mathcal{R}^{|\class{f}| \times |\class{f}|}$ is the attention weights matrix, and $\attkernel{q},\attkernel{k} \in \mathcal{R}^{d \times d}$ are trainable linear projection kernels.
Then, we normalize the weights and output 
\begin{equation*}
    \protos{'} =  \protos{f} +  \sigma (\att{\protos{f},\protos{f}} ) (\protos{f}\attkernel{v}),
\end{equation*}
where $ \sigma(\cdot)$ is a softmax function for each row in $\att{\protos{f},\protos{f}}$ and $\attkernel{v}\in \mathcal{R}^{d \times d}$ is another trainable linear projection kernel.
Then, we feed the average of $\protos{'}$ to a MLP function $\gen$ to get the negative prototype $\proto{}^-$.

\subsubsection{ATT-G} 
The above generators is suitable for \fsor problem. Now we consider the more challenging \gfsor task.  Directly employing the above methods may introduce bias towards $\class{*}$ as $\class{*}$ has plenty of training samples and the prototype $\protos{*}$ can be better-estimated compared against the few-shot prototypes $\protos{f}$. Hence we need another negative generator compatible with GFSOR , which should take care of both $\class{*}$ and $\class{f}$.
We build our ATT-G generator on top of a popular \gfsl method~\cite{gidaris2018dynamic}, which uses an attention mechanism to calibrate few-shot prototypes $\protos{f}$ with $\protos{*}$.
Specifically, we follow~\cite{gidaris2018dynamic, gidaris2019boosting, tian2020rethinking} to first   
train a network under large-scale classification task using the labeled samples of $\class{*}$ (\ie, pre-training) and use the weight in the last linear layer as many-shot class prototypes $\protos{*}$. 
Then we apply the attention block between $\protos{f}$ and $\protos{*}$ to generate the negative prototype $\proto{}^{-}$, i.e., 
\begin{equation}\label{eq:att_raw}
\att{\protos{f},\protos{*}} = \frac 1 {\sqrt{d}} (\protos{f}\attkernel{q} (\protos{*}\attkernel{k})^T),
\end{equation}
\begin{equation}\label{eq:att_agree}
    \protos{'} =  \protos{f} +  \sigma (\att{\protos{f},\protos{*}} ) (\protos{*}\attkernel{v}),
\end{equation}
and $\proto{}^{-}$ is similarly computed by feeding the average of $\protos{'}$ into a MLP $f_n$.
Furthermore, we'd like to filter out task-irrelevant information by applying a channel-wise gating mechanism on top of $\protos{f}$:
\begin{equation}
    \proto{c}' = \proto{c} \odot \phi(\gating(\frac{1}{N-1}\sum\nolimits_{i \in \class{f} \setminus \{c\}}\proto{i})),
\end{equation}
for $c \in \class{f}$ where $\odot$ and $\phi$ denote element-wise multiplication and sigmoid operation, and $\gating$ is a fully-connected layer.
Then, we use the updated $\protos{f'}$ to replace the input $\protos{f}$ in Eq.~\ref{eq:att_raw}.
Finally, we follow the order of many-shot prototypes $\protos{*}$, few-shot prototypes $\protos{f}$, and negative prototype $\proto{}^{-}$ to build the open-set classifier for a GFSOR task, where $\protos{*}$ are the weights in the last linear layer after pre-training.

\subsubsection{SEMAN-G} 
Inspired by recent cross-modal \fsl works \cite{li2020boosting,xing2019adaptive},  we further explore how class semantics could help model negative class. Specifically, we use a cross-modal attention mechanism on top of \textbf{ATT-G}. For each class $c \in \class{f} \cup \class{*}$, we concatenate $\proto{c}$ with its word embedding $\semanvec{c} \in \mathcal{R}^w$ along the channel to have $\semanvis{c} = [\proto{c},\semanvec{c}] \in \mathcal{R}^{w+d}$.
Then we use $\mathbf{Z}^f$ and $\mathbf{Z}^*$ instead of $\protos{f}, \protos{*}$ in Eq.~\ref{eq:att_raw} to calculate the attention. And stick to  Eq.~\ref{eq:att_agree} to take $\protos{*}$ as input since we are still comparing visual features for recognition.

\subsubsection{Multiple Negative Prototypes}
In addition, we can easily extend from single negative generation to multiple negative generation. Specifically, we can learn a set of generators $\{g_{n,i}\}_{i=1}^M$ to generate multiple negative prototypes for each task. For ATT, ATT-G, and SEMAN-G, to reduce the number of trainable parameters, we choose to share the linear projection kernels in attention mechanism used to calculate $\protos{'}$, but just train separate MLPs $\{f_{n,i}\}_{i=1}^M$ to synthesis multiple negative prototypes. In this way, we get multiple thresholds $\{\thre{a,i}\}_{i=1}^M$. Then the maximum threshold $\thre{a} = \text{max}\big(\{\thre{a,i}\}_{i=1}^M\big)$ is used as the final threshold for open-set recognition.

\subsection{Conjugate Training}

Here we present our conjugate training strategy towards (G)\fsor. Conjugate training is built upon the standard \fsor meta-training approach~\cite{Liu_2020_CVPR, snell2017prototypical}. We first go over the standard \fsor meta-training then introduce our method. 

\subsubsection{Standard FSOR Meta-Training}
Standard FSOR meta-training strategy~\cite{Liu_2020_CVPR,snell2017prototypical} trains the model by simulating \fsor tasks from the given \textit{base} dataset $\img{B}$. 
Specifically, it trains on a set of tasks sampled from the \textit{base} dataset $\img{B}$ where the images are from \textit{base} classes $\class{B}$. 
Within an $N$-way $K$-shot FSOR task $\task{}$, unknown sources are simulated using a different set of $N$ classes $\class{n}$, \ie, $\class{n} \subset \class{B}-\class{f}$ where $|\class{n}|=N$, and then randomly sample images belonging to $\class{n}$ from $\img{B}$ for $\query{}^n$. Then the model is trained using an objective, typically an open recognition loss within the sampled $\task{}$.
The standard \fsor meta-training can be generalized to \gfsor. For a GFSOR task $\task{*}$, we can sample $\class{f}$ and $\class{*}$ from $\class{B}$ and then simulate unknown sources as $\class{n} \subset \class{B}-(\class{f}\cup\class{*})$ where $|\class{n}|=N$. And a \gfsor training objective may be specified to learn the model for \gfsor.   
Note that, during inference time (i.e., meta-testing),  tasks are sampled from \textit{novel} dataset $\img{N}$ where the images are from classes $\class{N}$ and no sample from $\class{N}$ is seen during meta-training, \ie, $\class{B} \cup \class{N} = \varnothing$.

\subsubsection{Conjugate Tasks} 
The idea of conjugate training is to sample task pairs  whose few-shot examples of one task are used as the negative source of the other. Formally,
we define two tasks $\task{}_1= (\supp{1}, \query{1}^f, \query{1}^n | \class{f}_1)$ and $\task{}_2=(\supp{2}, \query{2}^f, \query{2}^n | \class{f}_2)$ as a \textit{conjugate task pair} when $\query{1}^n=\query{2}^f$ and $\query{1}^f=\query{2}^n$, \ie, the few-shot classes $\class{f}_1$ ($\class{f}_2$) in $\task{}_1$($\task{}_2$) is used as the negative source in task $\task{}_2$($\task{}_1$). For a conjugate GFSOR task pair ($\task{*}_1,\task{*}_2$), in addition, $\task{*}_1$ and $\task{*}_2$ share the same many-shot class $\class{*}$ and its queries $\query{}^*$. 

\subsubsection{Conjugate Training Loss} 
We use a  standard cross-entropy loss $\mathcal{L}_{CE}(\cdot,\cdot)$~\cite{murphy2012machine}. For an FSOR task $\task{}$, we use cosine similarity as $f_s$ and use $\query{}^f \cup \query{}^n$ to perform $(N+1)$-way classification. For each positive query $q \in \query{}^f$, we learn to maximize the class score of its label category by minimizing $\mathcal{L}_{CE}(y_q,q)$ where $y_q \in \{1,...,N\}$ is the class label of $q$. For each negative query $q \in \query{}^n$, we set its ground truth label as $N+1$ and maximize the threshold $\thre{a}$ by minimizing $\mathcal{L}_{CE}(N+1,q)$.
During conjugate training, we consider the dependency of negative sampling mentioned in~\cite{jeong2021few}. Without loss of generality, for a positive query $q \in \query{2}^f$  belonging to class $c_n \in \class{f}_2$ in $\task{}_2$, it is used as the negative query and is trained to have high similarity with the negative prototype in $\task{}_1$.

With a simple classification loss, the negative prototypes are optimized to learn a tight rejection boundary for a specific task.
Besides, for the attention-based generators, we also regularize the intermediate variables $\protos{'}$ as class-specific negative prototypes. For each prototype $\proto{c}' \in \protos{'}$ generated from a positive prototype $\proto{c}$, we can think of $\proto{c}'$ as the negative prototype for class $c$. Then, for each $\proto{c}'$, we minimize its similarity with queries of class $c$ and maximize its similarity of negative queries with a binary cross-entropy loss $\mathcal{L}_{BCE}$
\begin{align}
    \mathcal{L}_{neg}(c) &= \frac{1}{|\query{c,1}^f|}\sum\nolimits_{q\in \query{c,1}^f} \mathcal{L}_{BCE}(0,f_s(f(q),\proto{c}')) \nonumber \\
    &+ \frac{1}{|\query{1}^n|}\sum\nolimits_{q\in \query{1}^n} \mathcal{L}_{BCE}(1,f_s(f(q),\proto{c}')) \nonumber 
\end{align}
where $\query{c,1}^f= \{q|q \in \query{1}^f, y_q=c\}$ and $y_q$ denotes the class label of $q$.
Finally, without loss of generality, for $\task{}_1$ in the conjugate task pair ($\task{}_1, \task{}_2$), we have
\begin{equation}
    \mathcal{L}_{\task{}_1} = \mathcal{L}_{CE}(\query{1}^f \cup \query{1}^n) + \frac{1}{|\mathcal{C}_1^f|}\sum\nolimits_{c\in \mathcal{C}_1^f} \mathcal{L}_{neg}(c),
\end{equation}
and the total conjugate training loss is $\mathcal{L} = \mathcal{L}_{\task{}_1} + \mathcal{L}_{\task{}_2}$.

Similarly, for the network trained on \gfsor tasks, given a conjugate task pair ($\task{*}_1, \task{*}_2$), we have 
$$
\mathcal{L}_{\task{*}_1}^* = \mathcal{L}_{CE}(\query{1}^* \cup \query{1}^f \cup \query{1}^n) + \frac{1}{|\mathcal{C}_1^f|}\sum\nolimits_{c\in \mathcal{C}_1^f} \mathcal{L}_{neg}(c),
$$
where the class label for a negative query is $N+1+|\class{*}|$, and the total loss being $\mathcal{L}^* = \mathcal{L}_{\task{*}_1}^* + \mathcal{L}_{\task{*}_2}^*$. In this way, our \textit{conjugate training} involves the class-correlation during network training.

\section{Experiments and Analysis} \label{sec:experiment}
\subsubsection{Datasets} 

For \fsor tasks, we evaluate on two widely used public benchmarks: {MiniImageNet} \cite{vinyals2016matching}, {TieredImageNet} \cite{ren2018metalearning}. {MiniImageNet} \cite{vinyals2016matching} contains 100 classes and the class split for (meta-training, meta validation,meta-testing) is (64,16,20). Each class has 600 images. {TieredImageNet} \cite{ren2018metalearning} contains 608 classes and the class split is (351, 97, 160) while the \textit{base} dataset contains around 450K images. 
We evaluate \gfsor on {MiniImageNet} \cite{vinyals2016matching} and set the base classes during meta-training as the many-shot classes during meta-testing. We follow ~\cite{gidaris2018dynamic} and use another 300 images for each \textit{base} class for the \gfsor simulation. All images for the two datasets are sized to 84$\times$84. For \textbf{SEMAN-G}, we extract word embedding using GloVe~\cite{pennington2014glove}. More details of the datasets can be found in the supp. material. 

\begin{figure}
    \centering
    \includegraphics[width=\linewidth]{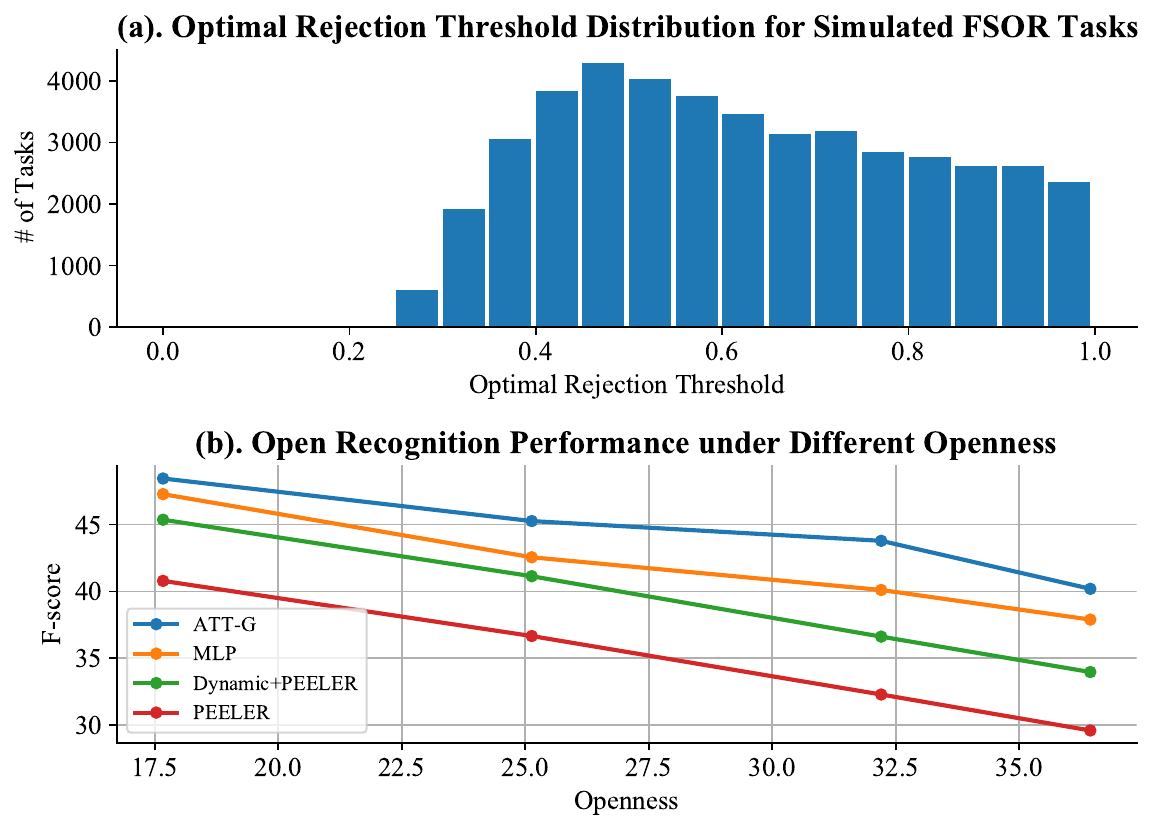}
    \caption{(a) Distribution of optimal rejection thresholds with Dynamic+PEELER on Mini-ImageNet. (b) Comparison of open recognition performance  under different openness.}
    \label{fig:hist_open}
\end{figure}

\begin{table}[t]
\setlength{\tabcolsep}{0.7em}
\caption{F1-score comparison on 5-way 1-shot FSOR tasks on Mini-ImageNet. $^*$: our implementation.}
\resizebox{\linewidth}{!}
{
\renewcommand{\arraystretch}{1.0}
\begin{tabular}{c|c|c|c|c}
\hlineB{3}
	 Neg. Gen. &  0.3 & 0.5 & 0.7 & 0.9 \\
	 \hline
	 PEELER & $~34.01~$  & $40.71 \spm 0.59$ & $~~41.44~~$ & $~~35.30~~$  \\ 
	 Dynamic+PEELER &  $~38.19~$ & $45.34 \spm 0.64$ & $~~44.10~~$ & $~~30.95~~$ \\ \hline \hline
	 \multicolumn{1}{c|}{Neg. Gen.} & \multicolumn{2}{c|}{Single Neg.} & \multicolumn{2}{c}{Multi Neg.} \\\hline
	 \multicolumn{1}{c|}{AVG}   &  \multicolumn{2}{c|}{${  45.6 \spm 0.71}$} &  \multicolumn{2}{c}{-}\\ 
	 \multicolumn{1}{c|}{MLP} & \multicolumn{2}{c|}{${46.12 \spm 0.74}$} &  \multicolumn{2}{c}{${ 47.21 \spm 0.72}$}\\ 
	 \multicolumn{1}{c|}{ATT} & \multicolumn{2}{c|}{${46.38 \spm 0.73}$} &  \multicolumn{2}{c}{$47.29{  \spm 0.70}$}\\ 
	 \multicolumn{1}{c|}{ATT-G} & \multicolumn{2}{c|}{${47.03 \spm 0.74}$} &  \multicolumn{2}{c}{ $48.19{  \spm 0.71}$}\\ 
	 \multicolumn{1}{c|}{SEMAN-G} & \multicolumn{2}{c|}{${47.95 \spm 0.72}$} &  \multicolumn{2}{c}{$50.10{  \spm 0.69}$}\\ 
\hlineB{3}
\end{tabular}
}
\label{tab:neg_gen}
\end{table}

\begin{table*}[t]
\setlength{\tabcolsep}{0.7em}
\caption{5-way 1-shot and 5-shot  FSOR results. $^*$: our implementation.}
\resizebox{\textwidth}{!}
{
\renewcommand{\arraystretch}{1.0}
\begin{tabular}{cc|c|c|c|c|c|c|c}
\hlineB{3}
	 \multirow{3}{*}{Algorithm} &  \multicolumn{4}{c|}{MiniImageNet,5-way} & \multicolumn{4}{c}{TieredImageNet,5-way} \\
	 \cline{2-9} 
	 &  \multicolumn{2}{c|}{1-shot} & \multicolumn{2}{c|}{5-shot} & \multicolumn{2}{c|}{1-shot} & \multicolumn{2}{c}{5-shot}  \\
	 \cline{2-9} 
	 &  Acc & AUROC & Acc & AUROC &Acc & AUROC & Acc & AUROC  \\
	 \hline
	 ProtoNet~\cite{snell2017prototypical} & $64.01\spm0.88$ & $51.81\spm0.93$ & $80.09$ & $60.39$ & $68.26$ & $60.73$ & $83.40$ & $64.96$  \\  
	 
	 FEAT~\cite{ye2020few} & $67.02\spm0.85$ & $57.01\spm0.84$ & $82.02$ & $63.18$ & $70.52$ & $63.54$ & $84.74$ & $70.74$ \\
	 \hline
	 
	 OpenMax~\cite{openmax2016} & $63.69\spm0.84$ & $62.64\spm0.80$ & $80.56$ & $62.27$ & $68.28$ & $60.13$ & $83.48$ & $65.51$ \\
	 
	 CounterFactual$^{*}$~\cite{Neal_2018_ECCV}  & $63.7{   \spm 0.83}$ & $64.17{   \spm 0.88}$  & $81.44{   \spm 0.54}$ & $71.58{   \spm 0.76}$ & $70.08{   \spm 0.94}$ & $71.04{   \spm 0.80}$  & $85.36{   \spm 0.60}$ & $78.66{   \spm 0.62}$   \\ 
	 \hline
	 
	 PEELER~\cite{Liu_2020_CVPR} & $65.86 \spm 0.85$ & $60.57 \spm 0.83$ & $80.61$ & $67.35$ & $69.51$ & $65.20$ & $84.10$ & $73.27$ \\ 
	 
	 SnaTCHer-T~\cite{jeong2021few} & $66.60\spm0.80$ & $70.17\spm0.88$ & $81.77$ & $76.66$ & $70.45$ & $74.84$ & $84.42$ & $82.03$ \\
	 SnaTCHer-L~\cite{jeong2021few} & $67.60\spm0.83$ & $69.40\spm0.92$ & $82.36$ & $76.15$ & $70.85$ & $\mathbf{74.95}$ & $85.23$ & $80.81$ \\
	 \hline
	 
    	 ATT (ours)  & $67.64{   \spm 0.81}$ & $71.35{ \spm 0.68}$  & $82.31{   \spm 0.49}$ & $79.85{   \spm 0.58}$ & $69.34{   \spm 0.95}$ & $72.74{   \spm 0.78}$  & $83.82{   \spm 0.63}$ & $78.66{   \spm 0.65}$   \\ 
	 
	 ATT-G (ours)  & $68.11{   \spm 0.81}$ & $72.41{   \spm 0.72}$  & $83.12{   \spm 0.48}$ & $79.85{   \spm 0.57}$ & $70.58{   \spm 0.93}$ & $73.43{   \spm 0.78}$  & $85.38{   \spm 0.61}$ & $81.64{   \spm 0.63}$   \\ 
	 
	SEMAN-G (ours)  & $\mathbf{68.24{   \spm 0.82}}$ & $\mathbf{72.85{   \spm 0.69}}$  & $\mathbf{83.48{   \spm 0.48}}$ & $\mathbf{82.07{   \spm 0.58}}$ & $\mathbf{71.06{   \spm 0.92}}$ & $74.27{   \spm 0.77}$  & $\mathbf{86.02{   \spm 0.58}}$ & $\mathbf{82.59{   \spm 0.57}}$   \\ 
	 
\hlineB{3}
\end{tabular}
}
\label{tab:fsor_sota}
\end{table*}

\begin{table*}[h]
\begin{center}
\setlength{\tabcolsep}{0.7em}
\caption{5-way generalized few-shot open-set recognition results on Mini-ImageNet.$^\dagger$: Implemented by CASTLE~\cite{CASTLE}. $^*$: our implementation.}
\resizebox{\textwidth}{!}
{\renewcommand{\arraystretch}{1.0}
\begin{tabular}{c cccc | cc | cc}
\hlineB{3}
\multirow{2}{*}{Algorithm} & \multicolumn{2}{c}{1-shot} & \multicolumn{2}{c|}{5-shots}   & 1-shot & 5-shots &  1-shot & 5-shots \\ \cline{2-9} 
& Arith. Mean & $\Delta$ & Arith. Mean & $\Delta$ & \multicolumn{2}{c|}{Harmonic Mean} & \multicolumn{2}{c}{AUROC} \\ \hlineB{1}
IFSL~\cite{ren2018incremental}    & $54.95 {\scriptstyle \pm 0.3}$ & $11.84$ & $63.04 {\scriptstyle \pm 0.3}$ & $10.66$	& -	& - & - & - \\					
L2ML~\cite{wang2017learning}    & $46.25 {\scriptstyle \pm 0.04}$ & $27.49$ & $45.81 {\scriptstyle \pm 0.03}$ & $35.53$ & $2.98 {\scriptstyle \pm 0.06}$ & $1.12 {\scriptstyle \pm 0.04}$ & -	& - \\	
ProtoNet2 $^\dagger$ &$53.93 {\scriptstyle \pm 0.08}$ & $22.09$ & $72.64 {\scriptstyle \pm0.08}$ & $11.41$ & $27.73{\scriptstyle \pm0.19}$ & $68.99	{\scriptstyle \pm 0.11}$ & - &	- \\	
CASTLE~\cite{CASTLE} & $66.48 {\scriptstyle \pm 0.11}$ & $9.94$ & $76.25 {\scriptstyle \pm 0.09}$ & $8.14$ & $64.29 {\scriptstyle \pm 0.14}$ & $\mathbf{75.79 {\scriptstyle \pm 0.1}}$ & - & - \\	
DynamicFSL$^*$~\cite{gidaris2018dynamic} & $60.85 {\scriptstyle \pm 0.14}$ & $12.97$ & $73.1 {\scriptstyle \pm 0.13}$ & $10.92$ & $60.13 {\scriptstyle \pm 0.13}$ & $69.8 {\scriptstyle \pm 0.09}$ & $67.56 {\scriptstyle \pm 0.17}$ & $72.86 {\scriptstyle \pm 0.11}$\\
\hline
\hline
ATT-G(ours) & $65.49 {\scriptstyle \pm 0.13}$ & 11.25 & $75.51 {\scriptstyle \pm 0.10}$ & 10.97 & $63.94 {\scriptstyle \pm 0.12}$  & $73.89 {\scriptstyle \pm 0.12}$ & $73.12 {\scriptstyle \pm 0.16}$ & $77.22 {\scriptstyle \pm 0.13}$ \\
SEMAN-G(ours) & $\mathbf{66.83 {\scriptstyle \pm 0.11}}$ & $10.24$ & $\mathbf{77.02 {\scriptstyle \pm 0.08}}$ & $9.78$ & $\mathbf{64.77 {\scriptstyle \pm 0.12}}$ & $75.62 {\scriptstyle \pm 0.09}$ & $\mathbf{73.55 {\scriptstyle \pm 0.14}}$ & $\mathbf{78.22{\scriptstyle \pm 0.11}}$\\
\hlineB{3}
\end{tabular}
}
\label{tab:gfsor}
\end{center}
\end{table*}

\subsubsection{Implementation Details} \label{sec:impl}

We use ResNet12 \cite{he2016deep} network as the feature backbone. Following \cite{gidaris2019boosting, tian2020rethinking}, we pre-train the ResNet12 and a classifier (a linear layer) with cross-entropy loss and a self-supervised rotation loss 
on the base set under fully-supervised classification task for $90$ epochs using a SGD optimizer with learning rate $0.05$ decayed by $10$ at epoch $60$. The weights in the linear layer are used as base-class many-shot prototypes $\protos{*}$  for \textbf{ATT-G} and \textbf{SEMAN-G}. Through the experiment, we interchangeably use the term \textit{base} and \textit{many-shot}.
During meta-training, the learning rate is set to $0.0001$ for the ResNet12 feature extractor, and $0.05$ for all other layers in the negative prototype generator.  The entire network is trained for $18k$ tasks with a SGD optimizer, where the learning rate is decayed when the validation accuracy saturates. 
During meta-testing, we follow \cite{jeong2021few, Liu_2020_CVPR} to randomly sample $600$ tasks, and report the average value with $95\%$ confidence interval for all the metrics. We use cosine similarity  \cite{gidaris2018dynamic} as the similarity function to compute per-class prediction scores.
For \fsor evaluation,  we follow \cite{Liu_2020_CVPR} to sample training and testing tasks, where we set $N=5$ and $K=1,5$. For each task, we sample $15$ positive queries from each few-shot class. For negative detection, we sample $5$ negative classes with each containing $15$ negative queries. 
For each \gfsor task, in addition to query samples from few-shot and negative classes, we select $75$ query samples for the \textit{base} classes (each class has at least one sample). Following the setup in~\cite{vinyals2016matching}, we randomly sample 1000 5-way \gfsor tasks to learn to generate an open-set classifier for the union of 64 base classes and 5 novel classes.

\subsubsection{Metrics}
To measure the standard closed-set classification performance, we report top-1 accuracy for \fsor tasks over few-shot classes.  For \gfsor, we follow the protocol defined in~\cite{ren2018incremental,ye2019learning}, and report both arithmetic mean and harmonic mean between mean accuracy of base samples and mean accuracy of novel samples. In addition, we report $\Delta$-value to measure the accuracy drop between prediction among specific classes (base or novel classes) and prediction among all combined classes, where a better classifier is supposed to balance the prediction and have low $\Delta$-value.
To measure the negative detection performance, we follow the protocol in \cite{Liu_2020_CVPR, jeong2021few} to report AUROC (area Under ROC Curve). 
To measure the overall open-set recognition performance, we follow the protocol in \cite{zhou2021learning} to report \textit{macro-averaged F1-scores} on all many-shot/few-shot and negative classes.

\subsection{\fsor Results}
\subsubsection{Comparison of Negative Generator}
We first compare different choices of negative generators on \fsor tasks in Tab.~\ref{tab:neg_gen}. Note that \textbf{ATT-G} and \textbf{SEMAN-G} can also be applied for \fsor and compared to other approaches since all models are trained using base set only (including base prototype) and doesn't use any extra data. We can see that attention-based methods are effective in negative generation as they are good at modeling inter-class relations. Adding class semantic information is also beneficial for discrimination.
Meanwhile, by enabling multiple negative prototypes, \ie, $M=5$, we can automatically estimate the threshold $\thre{a}$ with more flexibility, which  then achieve consistent performance gain in F1-score, when compared generating a single negative prototype $M=1$. For the following experiment results, we set $M=5$ for our methods. 

\subsubsection{Comparison with Threshold-based Classifier}
For threshold-based methods, threshold-tuning is crucial to get good recognition performance. To evaluate the overall open-set recognition, we compare macro-weighted F1-score. For threshold-based approaches, we define different thresholds and compute the corresponding F1-score.
We illustrate our result in both Tab.~\ref{tab:neg_gen} and Fig.~\ref{fig:hist_open}(a), where we consider two threshold-based classifiers PEELER\cite{Liu_2020_CVPR} and a combination of PEELER and our ATT-G method baseline, Dynamic\cite{gidaris2018dynamic}, which calibrates novel prototype with base-class prototypes. In detail, we apply PEELER's open-set training strategy on top of Dynamic.  

In Fig.~\ref{fig:hist_open}(a), we simulate 45k FSOR tasks and find their optimal rejection thresholds $\theta_m$ for the threshold-based approach Dynamic+PEELER. We plot the distribution of $\theta_m$, which shows that it covers a wide range between $0$ and $1$. It demonstrates that different \fsor tasks may need very different rejection threshold in practice with current threshold-based approach. And the overall recognition performance largely depends on threshold selection, as is shown in Tab.~\ref{tab:neg_gen}.
Our method instead automatically learn a task-adaptive rejection boundary, and we can see from Tab.~\ref{tab:neg_gen} that all our negative envision instantiations outperform threshold-based methods. Fig.~\ref{fig:hist_open}(b) further analyze the recognition behaviour under different openness~\cite{sun2020conditional}:

$$
\text{openness} = 1 - \sqrt{\frac {2|\class{f}|} {2|\class{f}| + |\class{n}|}}
$$ 
where we fix $|\class{f}| = 5$ and vary $|\class{n}|$ from $5$ to $15$. Similarly, we test for $600$ randomly selected \fsor tasks and take the average. As is validated by Fig.~\ref{fig:hist_open}, our method clearly outperforms threshold-based methods at all openness levels.

\subsubsection{Comparison with SOTA Methods}
We compare our method with other SOTA methods. The baselines we compare to include standard \fsl methods (ProtoNet, FEAT), large-scale OR methods (OpenMax, CounterFactual), and existing \fsor methods (PEELER, SnaTCHer). We cite most of the baseline results from \cite{jeong2021few}, and additionally compare to CounterFactual, a generative OR method which synthesize fake negative images and then train a $N+1$ classifier. To apply in our \fsor setting, we first train its GAN network on base set and use the support set to synthesize fake images. The averaged fake image feature is used as the negative prototype for \fsor. 

Tab.~\ref{tab:fsor_sota} demonstrates the results. Standard FSL methods perform poorly in negative detection due to its closed-set nature. Large-scale OR methods yields unsatisfactory performance especially on 1-shot classification. Interestingly, CounterFactual gives a relatively fair performance on negative detection, which also validates our concept of negative envision. But it's still much worse than our few-shot-specific negative generation strategy, which validates that our approach better suits for the limited data scenario. 
Both \textbf{ATT-G} and \textbf{SEMAN-G}  outperform other methods on Mini-ImageNet and get comparable result on Tiered-ImageNet.

\begin{table}[t]
\begin{center}
\caption{Ablate study on conjugate training.  We report 5-way-1-shot \fsor result using ATT-G over three metrics on both datasets.}
\resizebox{\linewidth}{!}
{
    \begin{tabular}{cc|c|c}
    \hlineB{3}
    {Dataset} & {Metric} & {w/o Conjugate}  &{w/ Conjugate} \\ \cline{1-4} 
     &  ACC  &         $66.28 \spm 0.84$ &  $68.11{   \spm 0.81} $\\
     Mini  & AUROC &       $71.80 \spm 0.77$  & $72.41{ \spm 0.72} $\\
     ImageNet &  F1-score  &    $46.94 \spm 0.68$         &              $48.19 \spm 0.71 $ \\\cline{1-4} 
      &  ACC  &        $70.08 \spm 0.94$         &                     $70.58 \spm 0.93$ \\
    Tiered & AUROC &      $71.84 \spm 0.82$         &                   $73.43 \spm 0.78$ \\
    ImageNet&  F1-score  &    $50.23 \spm 0.77$         &                  $51.56 \spm 0.81$   \\\cline{1-4} 
    \hlineB{3}
    \end{tabular}
}
\label{tab:fsor_ablate}
\end{center}
\end{table}
\subsubsection{Ablation Study on Conjugate Training}

Tab.~\ref{tab:fsor_ablate} shows the impact of conjugate training. We observe consistent improvement on all metrics and datasets, validating that conjugate training efficiently boosts the learning process by enabling mutual supervision from two tasks.

\subsection{\gfsor Results}
In Tab.~\ref{tab:gfsor}, we compare \textbf{ATT-G} and \textbf{SEMAN-G} with other standard methods on \gfsor tasks.
Under the more challenging GFSOR setting, we achieve comparable \gfsl classification accuracy with SOTA method and significantly improve the AUROC score which measures negative query detection. 
In addition, as  \gfsl methods are not trained to envision negative prototype but has more classes to recognize during evaluation, it will be challenging to manually set a threshold to reject negative queries while maintaining high classification accuracy. Thus, it is necessary to learn to dynamically generate threshold for each query for \gfsor.

\subsection{More Experiments}

We further conduct \fsor experiments on two few-shot benchmark datasets: {CIFAR-FS}\cite{bertinetto2018metalearning}, {FC100} \cite{NEURIPS2018_66808e32}.
CIFAR-FS~\cite{bertinetto2018metalearning} contains 100 classes with the class split for (64,16,20).
FC100~\cite{NEURIPS2018_66808e32} contains 100 classes with the class split (60,20,20). 
Each class has 600 images and all images for the two sets are of size is 32$\times$32.
As shown in Tab.~\ref{tab:fsor_cifar} and \ref{tab:fsor_fc100}, we compare our methods with the threshold-based methods and direct application of large-scale open-set recognition methods. Consistent with Tab.~\ref{tab:fsor_sota}, for low-resolution dataset, our method achieves the best performance on both classification accuracy and negative query rejection, which again demonstrates the effectiveness of our approach.

\begin{table}[t]
\setlength{\tabcolsep}{0.7em}
\caption{5-way FSOR results CIFAR-FS.$^*$: our implementation. } %
\resizebox{\linewidth}{!}
{
\renewcommand{\arraystretch}{1.0}
\begin{tabular}{cc|c|c|c}
\hlineB{3}
	 \multirow{2}{*}{Algorithm} & \multicolumn{2}{c|}{1-shot} & \multicolumn{2}{c}{5-shot} \\ \cline{2-5}
	 &  Acc & AUROC & Acc & AUROC \\
	 \hline
	 OpenMax\cite{openmax2016} & $71.65$ & $50.21$ & $85.66 $ & $75.78 $ \\
	 CounterFactural\cite{Neal_2018_ECCV} & $71.71$ & $72.57$ & $85.71$ & $80.44$ \\
	 PEELER\cite{Liu_2020_CVPR}$^*$ & ${71.47}$  & ${71.28}$ & ${85.46 }$  & $75.97 $ \\
	 Dynamic\cite{gidaris2018dynamic} & $71.56$ & $66.89$ & $85.78$ & $76.03$ \\
	 ATT-G (ours)  & $72.43{ }$ & $76.72{}$  & $86.52$ & $84.64$  \\ 
	SEMAN-G (ours)  & $74.55{ }$ & $78.10{}$  & $86.71$ & $86.47$ \\ 
	 
\hlineB{3}
\end{tabular}
}
\label{tab:fsor_cifar}
\end{table}

\begin{table}[t]
\setlength{\tabcolsep}{0.7em}
\caption{5-way FSOR results on FC100. $^*$: our implementation.} %
\resizebox{\linewidth}{!}
{
\renewcommand{\arraystretch}{1.0}
\begin{tabular}{cc|c|c|c}
\hlineB{3}
	 \multirow{2}{*}{Algorithm} & \multicolumn{2}{c|}{1-shot} & \multicolumn{2}{c}{5-shot} \\ \cline{2-5}
	 &  Acc & AUROC & Acc & AUROC \\
	 \hline
	 OpenMax\cite{openmax2016} & $44.70$ & $50.10$ & $60.11$  & $57.78$ \\
	 CounterFactural\cite{Neal_2018_ECCV} & $44.53$ & $57.20$ & $61.12$ &  $62.35$ \\
	 PEELER\cite{Liu_2020_CVPR}$^*$ & ${44.45}$ & ${55.86 }$ & ${60.86 }$ &  ${61.07}$ \\
	 Dynamic\cite{gidaris2018dynamic} & $44.88$ & $55.62$ & $60.45 $ & $59.01 $ \\
	 ATT-G (ours)  & $45.11$ & $59.55 $  & $61.18$ & $63.34 $  \\ 
	SEMAN-G (ours)  & $46.01$ & $59.73 $  & $62.18$ & $64.46 $ \\ 
	 
\hlineB{3}
\end{tabular}
}
\label{tab:fsor_fc100}
\end{table}

\section{Conclusion}
In this work, we show the limitation of threshold-based approaches for few-shot open-set recognition where different tasks may need very different rejection threshold and hence the tuning process could be challenging. To this end, we propose our task-adaptive negative envision approach towards (G)\fsor, where negative prototypes are computed from few/many-shot class examples. We study the different design of negative generator, and find attention-based generator works the best; adding class semantics further improves the performance. 
We also introduce a new conjugate class training strategy to better facilitate the learning process. Extensive experiments demonstrate  the effectiveness of our approach.
We note the limitation where we assume the negative source only being the images from other categories. Other possible negative sources include, e.g., data from different domains, adversarial data, etc. We will leave those as future work and study they affect our approach. 

\section{Acknowledgement}
This research is based upon work supported by the Intelligence Advanced Research Projects Activity (IARPA) via Department of Interior/Interior Business Center (DOI/IBC) contract number D17PC00345. The U.S. Government is authorized to reproduce and distribute reprints for Governmental purposes not withstanding any copyright annotation thereon. Disclaimer: The views and conclusions contained herein are those of the authors and should not be interpreted as necessarily representing the official policies or endorsements, either expressed or implied of IARPA, DOI/IBC or the U.S. Government.

\newpage
{\small
\bibliographystyle{ieee_fullname}
\bibliography{egbib}
}

\end{document}